\pdfoutput=1

\documentclass[11pt]{article}

\usepackage[final]{acl}

\usepackage{times}
\usepackage{latexsym}

\usepackage[T1]{fontenc}

\usepackage[utf8]{inputenc}

\usepackage{microtype}

\usepackage{inconsolata}

\usepackage{graphicx}

\usepackage{xcolor}
\usepackage{soul}
\usepackage{tabularx}
\usepackage{makecell} 
\usepackage{booktabs}
\usepackage{titlesec}
\usepackage[htt]{hyphenat}
\usepackage{longtable}
\usepackage{diagbox}
\usepackage{comment}
\usepackage{enumitem}
\usepackage{appendix}
\usepackage{multirow}

\definecolor{clarity_color}{RGB}{178, 229, 206}
\definecolor{hatespeech_color}{RGB}{255, 179, 163}
\definecolor{fear_color}{RGB}{173, 216, 230}
\definecolor{alternative_color}{RGB}{255, 239, 153}
\definecolor{narrative_color}{RGB}{240, 200, 220}
\definecolor{critical_color}{RGB}{230, 230, 250}
\definecolor{fact_color}{RGB}{255, 200, 100}
\newcommand{\hlclarity}[1]{{\sethlcolor{clarity_color}\hl{#1}}}
\newcommand{\hlhatespeech}[1]{{\sethlcolor{hatespeech_color}\hl{#1}}}
\newcommand{\hlfact}[1]{{\sethlcolor{fact_color}\hl{#1}}}
\newcommand{\hlalternative}[1]{{\sethlcolor{alternative_color}\hl{#1}}}
\newcommand{\hlnarrative}[1]{{\sethlcolor{narrative_color}\hl{#1}}}
\newcommand{\hlcritical}[1]{{\sethlcolor{critical_color}\hl{#1}}}
\newcommand{\hlfear}[1]{{\sethlcolor{fear_color}\hl{#1}}}

\title{Debunking with Dialogue? Exploring AI-Generated Counterspeech to Challenge Conspiracy Theories}

\author{Mareike Lisker \and Christina Gottschalk \and Helena Mihaljević \\
         University of Applied Sciences (HTW) Berlin, Germany \\\texttt{\{lisker,mihalje\}@htw-berlin.de} 
         }

\begin{document}
\maketitle
\begin{abstract}
Counterspeech is a key strategy against harmful online content, but scaling expert-driven efforts is challenging. Large Language Models (LLMs) present a potential solution, though their use in countering conspiracy theories is under-researched. Unlike for hate speech, no datasets exist that pair conspiracy theory comments with expert-crafted counterspeech. We address this gap by evaluating the ability of GPT-4o, Llama 3, and Mistral to effectively apply counterspeech strategies derived from psychological research provided through structured prompts. Our results show that the models often generate generic, repetitive, or superficial results. Additionally, they over-acknowledge fear and frequently make up facts, sources, or figures, making their prompt-based use in practical applications problematic. 
\end{abstract}

\section{Introduction}
\textit{Conspiracy theories} (CTs) are ubiquitous, often emerging as attempts to identify underlying causes of unexpected or significant events. The resulting narratives allege clandestine machinations by powerful entities perceived as malevolent. CTs are emotionally charged and compelling, forming closed belief systems characterized by internal coherence, which makes them difficult to refute \cite{lepoutre_narrative_2024}. They are often associated with harmful social, health and political consequences \cite{douglas_understanding_2019}. For example, CTs can lead individuals to reject scientific consensus \cite{weigmann_genesis_2018} and influence societal attitudes about critical issues, such as climate change or vaccine policies \cite{jolley_social_2014}. Belief in CTs can undermine trust in democratic institutions, amplify social tensions and lead to violence \cite{vergani_hate_2022}.

\textit{Counterspeech} (CS) has emerged as an important strategy for mitigating the impact of harmful content in the online sphere. It describes communicative measures designed to refute undesirable content while respecting the principle of freedom of expression \cite{schieb_governing_2016}. Many non-governmental organizations (NGOs) employ it as a strategy to combat hate speech (HS), misinformation and CTs \cite{rathje_no_2015,unesco_addressing_2022}. While shifting the beliefs of an ideologically committed user who is propagating CTs might be rather difficult \cite{kreko_countering_2020}, counterspeech can positively affect the discourse norms of the audience \cite{bojarska_dynamics_2018}.

Manual moderation for comprehensive counterspeech is challenging due to high volumes and the rapidly evolving social media landscape \cite{bonaldi_nlp_2024}. In this context, supporting NGOs by automatically generating counterspeech that requires little post-processing becomes increasingly vital \cite{mun_counterspeakers_2024}. Large Language Models (LLMs) could potentially assist \cite{chung_conan_2019}. 
While research has explored using LLMs for counterspeech against HS, there is a notable gap in addressing CTs. While CTs can incite hatred and often perpetuate antisemitic or racist narratives \cite{samantha_hay_alternative_2020}, they typically differ linguistically from HS.

This research gap motivates our study, which investigates the potential of LLMs to generate counterspeech for CTs. We focus on the scenario of an NGO employee leveraging LLM-generated suggestions to respond to comments promoting CTs. 
Since no expert-based datasets containing examples of counterspeech to CTs could be identified, we rely on zero-shot prompting to evaluate the LLMs. Our central research question is thus:\\
\textbf{(RQ)} To what extent can current LLMs guided by prompt-based strategy instructions generate effective counterspeech against conspiracy theories?

The prompts are based on four strategies drawn from the literature,
particularly in psychology: \hlfact{fact-check based refutation}, \hlalternative{providing alternative explanations}, \hlnarrative{storytelling of a counter-narrative}, and \hlcritical{encouraging critical thinking}. These strategies are considered promising for countering CT beliefs in audiences not yet fully absorbed by them. 
We applied three LLMs---GPT-4o, Llama 3 (8B), and Mistral (7B)---to a dataset of 152 comments from the platform X that promote two CT themes: (1) narratives involving the `deep state', `NWO' and `globalists', and (2) claims related to `geo- and bioengineering'. The generated CS is qualitatively evaluated according to the established criteria. In addition to assessing individual comments, we analyze the lexical and semantic diversity of the generated CS to evaluate its practical utility for civil society actors.

Our findings show that all three models struggled to produce CS with substantial depth, often defaulting to superficial statements and avoiding narrative storytelling as a strategy. Responses frequently contained generic and repetitive phrases, resulting in limited linguistic variety. Notably, approximately 10\% of outputs contained confabulations which were often difficult to spot. Differences between the models were low and mostly not significant.
These findings suggest that current LLMs are not yet effective for the generation of CS to CTs in a zero-shot setting, underscoring the need for further research on grounding LLMs in relevant domain knowledge.

\section{Related Work}
\label{sec:background}

Research on automated CS generation has gained traction, but has primarily focused on HS. To our knowledge, only \cite{costello_durably_2024} have examined the use of LLMs to generate CS for CTs. In lab experiments, participants engaged in multi-turn dialogues with GPT-4 Turbo, resulting in a statistically significant reduction in self-reported CT belief. This suggests general potential for CT belief revision through LLM interaction, though in a direct, dialogue-based setting differing from the social media context.

We found no publicly available datasets containing examples of effective CS to CTs communicated online. Existing datasets on online CTs (e.\,g., \cite{langguth_coco_2023,steffen_codes_2023}) were mainly created for classification tasks \cite{liu_conspemollm_2024,peskine_detecting_2023,pustet_detection_2024}, and are mostly related to COVID-19.

In the remainder, we focus on LLM-based CS generation in the context of HS, as we found no prior work addressing these aspects for CTs.

\label{sec:background-automated}
The majority of approaches in CS generation rely on fine-tuning methods \cite{qian_benchmark_2019,tekiroglu_using_2022}, using datasets from platforms such as Reddit and Gab \cite{qian_benchmark_2019}, compiled by NGO operators \cite{chung_conan_2019}, or generated through models \cite{bonaldi_human-machine_2022}. More recently, zero-shot and few-shot prompting of various GPT-based models has shown promising results for generating CS against HS \cite{chung_towards_2021,ashida_towards_2022,zheng_what_2023,halim_wokegpt_2023}. Some studies have explored tailored approaches, such as using type-prompts to specify predefined styles like humor or facts \cite{saha_zero-shot_2024} or personalizing CS based on author profiling (e.\,g., age and gender) \cite{doganc_generic_2023} to fit specific contexts. Across these studies, the newer GPT models demonstrate strong performance.

Some studies have integrated automated CS generation into interactive tools. For instance, generation of contextualized CS was embedded within a content monitoring tool designed to address GDPR and DSA compliance \cite{fillies_hate_2024}, or incorporated into a dashboard enabling users to semi-automatically generate CS using text and memes \cite{smedt_feast_2021}. 

Bonaldi et al. present a comprehensive guide to using NLP to generate CS. They emphasize that while research in this area is growing, there is currently a lack of standardized definitions and best practices to integrate the different approaches \cite{bonaldi_nlp_2024}.


Ashida and Komachi highlight that evaluating CS is challenging due to the lack of established standards \cite{ashida_towards_2022}. Some studies use metrics like BLEU, GLEU, or METEOR to compare generated text to a `ground truth'. However, without CS references, these metrics are not applicable. Moreover, unlike machine translation, effective CS may include a wide range of diverse responses, making such metrics less suitable. 
Nonetheless, these measures can assess the diversity and novelty of outputs \cite{saha_zero-shot_2024}. Furthermore, classification models can be utilized to evaluate sentiment or argument characteristics \cite{saha_zero-shot_2024}.

In addition to quantitative methods, most researchers rely on human annotators for more in-depth evaluations with regard to `informativeness' and `offensiveness' \cite{ashida_towards_2022}, or `suitableness', and `intra-coherence' \cite{chung_towards_2021}, `grammaticality', `specificity' \cite{tekiroglu_using_2022}, or `diversity', `relevance' and `language quality' \cite{zhu_generate_2021}. 

\section{Counterspeech Against Conspiracy Theories}

Conspiracy theories
are interpretative frameworks attempting to explain events as the covert actions of powerful, malicious groups or individuals pursuing self-interest at public expense \cite{aaronovitch_voodoo_2010, byford_conspiracy_2011, keeley_conspiracy_1999}, overlooking actual, more intricate causes \cite{popper_offene_2003}. 
They thus usually need actors (e.\,g., `globalists') with a malicious goal (e.\,g., destabilizing national economies) through an action or strategy (e.\,g., manipulating weather) carried out in secrecy \cite{samory_conspiracies_2018}. On social media, these narratives often appear in fragmented forms, with some components omitted or implied \cite{steffen_codes_2023}. 
While CTs can overlap with HS \cite{baider_accountability_2023}, they have their own unique characteristics with regard to function and linguistic manifestation \cite{samantha_hay_alternative_2020}.


Interventions against CTs can be categorized as preventive (pre) or harm-reducing (post), as well as targeting the sender or the audience \cite{kreko_countering_2020}. As individuals deeply invested in conspiracy beliefs are particularly resistant to change \cite{kreko_countering_2020},
we focus on addressing the recipients of CTs, or `cyber-bystanders', in a social media context, post-exposure to a CT.

Study findings show that ``there is no such thing as perfectly refuted misinformation'' \cite[p.\,4]{kreko_countering_2020}, since part of the `essence' of the conspiracy remains embedded in the mental model of an individual. Nevertheless, research suggests that the design and formulation of CS can influence its impact \cite{chung_understanding_2023}.

\section{Data and Methods}

\subsection{Dataset}

As we target an audience not (yet) fully absorbed by conspiracy narratives, we sourced data from X (formerly Twitter), where exposure to CT content may reach broader, more diverse audiences. 
Our focus was on messages containing key elements of conspiracy narratives, such as identifiable actors, strategies, or goals, ensuring that the intended audience would likely grasp the central idea of the CT. Additionally, we prioritized messages with at least 200 likes or replies, as these are more likely to represent content with a wide reach.

The data was collected using keywords associated with two current and rather well-known CT themes, which can be broadly characterized as hate-based and fear-driven, namely theories revolving around `deep state', `NWO' and `globalists' on the hate-based side, and those relating to `geo- and bioengineering' on the fear-based side. The two sets of keywords (see Table \ref{tab:themes_keywords}) were iteratively enriched following an exploration of conspiratorial posts and hashtags on X. Note that we did not aim to create an exhaustive set of keywords, as we are mainly interested in identifying texts that can be clearly categorized as conspiratorial by an LLM. In total, we collected 152 posts (76 per category). All data and methods are available in the project repository.\footnote{\url{https://github.com/HTW-Social-Data-Science/Debunking_with_Dialogue}}

\begin{table}[ht]
\centering
\small
\caption{Keywords used to collect CT messages.}
\label{tab:themes_keywords}
\begin{tabular}{@{}ll@{}}
\toprule
\textbf{CT Theme} & \textbf{Keywords} \\
\midrule
Hate-based & \begin{tabular}[c]{@{}l@{}}deepstate, NWO, WW3, QAnon, globalists\end{tabular} \\ \hline
Fear-driven & \begin{tabular}[c]{@{}l@{}}geo-engineered, Big Pharma, bio-engineered,\\ HAARP, chemtrails, vaccine RFK Jr.\end{tabular} \\
\bottomrule
\end{tabular}
\end{table}


\subsection{Model Overview}
We prompted \textit{GPT-4o} \cite{openai_gpt-4o_2024} and two open-source models, \textit{Mistral} (\texttt{Mistral-7B-Instruct-v0.3}, \cite{mistral_mistral_2024}) and \textit{Llama 3} (\texttt{Meta-Llama-3-8B-Instruct}, \cite{aimeta_llama_2024}). In multiple previous studies, GPT-4o outperformed other models (cf., e.\,g. \cite{doganc_generic_2023,ashida_towards_2022}), and thus serves as a high-performance but also closed-source and cost-intensive solution. The smaller open models were selected for their accessibility and lower operational costs, making them more viable for practical applications in civic society settings. 

We accessed the open models via Hugging Face and GPT-4o through OpenAI's API. We set \texttt{temperature} and \texttt{top\_p} according to default or recommended values provided in the respective model documentations, ensuring optimal performance under consistent conditions.\footnote{For Llama 3, \texttt{temperature} was set to 0.6 and \texttt{top\_p} to 0.9, while both hyperparameters were set to 1 for the other two models.} All experiments were conducted in October and November 2024.

\subsection{Prompt Design}
\label{sec:prompt}
We opted for zero-shot prompting, including role and style prompting, as well as step-by-step instructions, as recommended by LLM providers and recent research (cf., e.\,g., \cite{schulhoff_prompt_2024}). The instructions primarily reflect four strategies for countering CTs that, according to 
current literature, are likely to enhance the effectiveness of CS: \hlfact{fact-check based refutation (\textit{Fact})}, \hlalternative{providing alternative explanations (\textit{Alt})}, \hlnarrative{storytelling of a counter-narrative (\textit{Narr})}, and \hlcritical{encouraging critical thinking (\textit{Crit})}. While \hlfact{\textit{Fact}} involves correcting misinformation through fact-based rebuttals, \hlalternative{\textit{Alt}} emphasizes providing more holistic alternative explanations. \hlnarrative{\textit{Narr}} uses narrative techniques to engage the audience emotionally, whereas \hlcritical{\textit{Crit}} aims at revealing the manipulative techniques used in conspiracy theories and encouraging reflective questioning.

In the following, we illustrate the different counterspeech strategies by sample answers to an example CT. For the theoretical underpinning of the prompt design, see Appendix \ref{app:strategies_literature}.

CT: ``COVID-19 was created as a bio-weapon in a lab to control the population.''
\\
\hlfact{\textit{Fact}}: ``Independent scientific bodies, including the World Health Organization and international virology experts, have found no evidence that COVID-19 was engineered or intentionally released, as noted in a 2020 article by Kristian G. Andersen in Nature Medicine. Genetic analyses indicate natural evolution, and similar coronaviruses found in animals, especially bats, support a zoonotic origin as stated in Wikipedia.''
\\
\hlalternative{\textit{Alt}}: ``Viruses like COVID-19 have emerged before through natural processes. Scientists have found that the virus shares much of its genetic code with coronaviruses found in bats and other animals, suggesting it likely jumped from animals to humans—something that has happened before with SARS and MERS. In a world where people and animals are increasingly in close contact, and global travel is constant, it's more likely that this virus emerged through natural transmission rather than being engineered.''
\\
\hlnarrative{\textit{Narr}}: ``In the 1990s, Dr. Lam discovered a virus that had jumped from bats to humans in Malaysia. When COVID-19 emerged, some suspected a lab origin, but Dr. Lam knew that the genetic code of the virus told a different story: it closely matched viruses found in bats, showing a natural origin. He explained that as humans encroach on wildlife habitats, viruses can spill over, and that this is what likely happened with COVID-19.''
\\
\hlcritical{\textit{Crit}}: ``Mhh, but if COVID-19 was a bio-weapon designed for control, why would it spread uncontrollably across all countries, including those allegedly behind it? Would a government really create something that harms its own economy, overwhelms its hospitals, and puts its own leaders at risk?'' 

The prompt formulation  was iteratively adjusted using a sample of messages and GPT-4o. 
The limit of 200 token was adjusted based on experiments with Llama 3 and Mistral as these had more difficulties with restricting to a smaller limit. See Appendix \ref{app:prompt} for the final version of the prompt.

\subsection{Counterspeech Annotation Framework}
Since comparing model output with `correct' counterspeech through quantitative metrics such as BLEU or METEOR cannot be employed, we rely on manual annotation of model outputs based on the prompt instructions.

We annotated a total of 456 responses across 12 criteria, each reflecting a specific aspect of the prompt instructions. The criteria are color-coded to correspond to the respective step in the prompt (see Appendix \ref{app:prompt}) as well as the respective literature they were derived from (see \ref{app:strategies_literature}):

\begin{table}[ht]
\small
\caption{Annotation criteria by prompt step and counterspeech strategy.}
\label{tab:annotation_criteria}
\centering
\begin{tabular}{@{}p{2.3cm}p{5.1cm}@{}}
\toprule
\textbf{Step / Strategy} & \textbf{Annotation Criteria} \\
\midrule
\multirow{3}{=}{\hlclarity{Clarity \& Restraint}} 
  & Conciseness and clarity \\
  & Repetition of false or harmful content \\
  & Use of stigmatizing language \\
\midrule
\multirow{3}{=}{\hlhatespeech{Hate Speech}} 
  & Hate speech identification\\
  & Hate speech condemnation\\
  & CT avoidance in hateful context \\
  \midrule
\multirow{2}{=}{\hlfear{Fear \& Empathy}} 
  & Fear identification \\
  & Appropriate empathy for fear \\
\midrule
\multirow{4}{=}{Strategies}
  & \hlfact{\textit{Fact}: Refute based on fact-checks} \\
  & \hlalternative{\textit{Alt}: Provide alternative explanations} \\
  & \hlnarrative{\textit{Narr}: Counter conspiracy with narrative} \\
  & \hlcritical{\textit{Crit}: Encourage critical thinking} \\
\bottomrule
\end{tabular}
\end{table}

Three annotation criteria were treated as binary variables: \textit{Hate speech identification}, \textit{Fear identification}, and \textit{Use of stigmatizing language}. The first two were coded based on a comparison between the original message and the model output. The latter indicates the presence or absence of the terms `conspiracy (theory)', `misinformation', `debunk-' and `unfounded'. The remaining 9 criteria were evaluated using Likert scales, which are available in the project repository.

To evaluate the annotation schema, a random sample of 50 responses---presented without any information about the respective models---was annotated independently by two authors. The inter-annotator agreement, measured using Krippendorf's Alpha, ranged from 0.35 (\textit{Conciseness and clarity}) to 0.83 (\textit{Hate speech condemnation} and \textit{Appropriate empathy for fear}). Disagreements were discussed and resolved collaboratively, and the annotation schema was refined accordingly. The lower agreement in case of (\textit{Conciseness and clarity}) was due to the initially underspecified criterion. In the subsequent discussion, we defined what to consider as `non-clear' or `non-concise', such as an unnecessary intro or outro, excessive mentioning of the CT, or patronizing formulations such as ``It's important to keep in mind ...''. The remaining data was split equally and annotated by one author each in close communication, with $\sim\!\!25\%$ of the records discussed jointly. Likert scales were iteratively refined, and prior annotations were retrospectively updated to ensure consistency. 

\subsection{Comparison of Models}
We quantitatively compared how the models implemented CS strategies based on the annotated Likert scale ratings. The only post-processing applied was separating the meta report from the actual CS via the \texttt{<XXX>}-tag. Due to the non-normal distribution of the values, we used the non-parametric Friedman test ($\alpha= 0.05$) to compare all three models. For significant results, we applied the the post-hoc Wilcoxon Signed-Rank test to identify model pairs with significant differences. Effect sizes were measured using Kendall's W. Comparisons were limited to records with non-missing values for all models. Bonferroni correction was applied to adjust p-values and control the family-wise error rate.

Furthermore, we assessed the response diversity using various lexical and semantic measures, as repetitive outputs would be less useful in the practical setting guiding our study.

\section{Results}
We first evaluate how well the models followed core prompt instructions related to structure, tone, and issue-specific handling.

\subsection{Adherence to Prompt Instructions}
\label{sec:results_explicit_instructions}

\paragraph{Meta Report and Length Restriction.}
The models showed varying adherence to simple prompt instructions, such as listing how hate speech was handled and which strategies were applied (the `meta report'), separating this information from the CS using the \texttt{<XXX>} tag, and keeping the CS under 200 tokens. 
GPT-4o consistently included the meta report, while Llama 3 and Mistral omitted it in 26–28\% of outputs (see col. 1 in Table \ref{tab:counterspeech_compliance}). The meta report in GPT-4o and Llama 3 responses was short, accounting for $\sim 1\%$ and $\sim 4\%$ of the output, respectively, while Mistral produced substantially longer explanations ($>20\%$ on average). GPT-4o correctly separated the meta report from the generated CS in over 96\% of outputs, enabling easy rule-based parsing. In contrast, Llama 3 responses required manual post-processing, i.\,e., separating the meta report from the CS, in 55\% of cases and Mistral failed to use the tag \texttt{<XXX>}, necessitating manual post-processing for all outputs. 
Mistral reported on HS handling in only 25\% of outputs, compared to 56\% for GPT-4o and 68\% for Llama 3 (see col. 3 in Table \ref{tab:counterspeech_compliance}). Interestingly, all models frequently misinterpreted the requirement to acknowledge fear---when present---as a CS strategy, often labeling it as `Strategy 5,' thus indicating confusion about the prompt instructions.

\begin{table}[ht] 
\centering 
\small
\caption{Compliance with prompt instructions: (1) inclusion of a meta report, (2) correct CS separation, (3) HS handling specified, and (4) mean CS length.
}
\label{tab:counterspeech_compliance}
\begin{tabularx}{\columnwidth}{l|p{1.1cm}p{1.1cm}p{1.2cm}X}
\toprule
model & Meta report (\%) & CS sep. (\%) & HS handling (\%) & Mean length \\
\midrule
GPT-4o & 100.00 & 96.05 & 55.92 & 709.38\\
Llama 3 & 71.71 & 55.05 & 68.22 & 1267.41\\
Mistral & 74.34 & 0.00 & 25.66 & 794.74\\
\bottomrule
\end{tabularx}
\end{table}

While Mistral and GPT-4o achieved an average character length of less than 800, thus mostly obeying the instruction to stay below 200 token, Llama 3 responses exceeded the limit with more than 1200 characters on average (see col. 4 in Table \ref{tab:counterspeech_compliance}).

\paragraph*{\hlclarity{Clarity \& Restraint.}}
GPT-4o significantly outperformed Llama 3 in clarity and conciseness, scoring 3.48 versus 3.12 on the Likert scale (effect size of 0.62). Mistral, with a mean clarity score of 3.36 lying in between, showed the greatest variability (SD=1.1) in performance.

GPT-4o showed the highest compliance with instructions to avoid terms like `conspiracy theory' and `misinformation,' using them in only 7.23\% of outputs, compared to 38.15\% for Mistral and 79.6\% for Llama 3. It also adhered more closely to the rule against repeating harmful content, with an average Likert score of 3.78, significantly outperforming Mistral (2.82) and Llama 3 (2.48). Mistral and Llama 3 frequently reinforced harmful content by including hashtags like `\#globalists', `\#DeepStateAgenda' or `\#QAnon', effectively promoting the narratives they were meant to counter.
Notably, a score of 4 is already considered high, as effectively refuting false claims without repeating any of them poses a significant challenge---even for humans.

\paragraph*{\hlhatespeech{Hate Speech.}}
The dataset included 17 messages containing explicit or implicit HS that the models were supposed to recognize and condemn without engaging with the CT content. GPT-4o and Llama\,3 achieved F1 scores of 0.75 and 0.69, respectively, in detection, while Mistral scored only 0.30 due to poor recall (differences not significant per McNemar's test). Overall condemnation of HS yielded mean Likert scores between 1.18 (Mistral) and 2.1 (Llama 3), reflecting limited effectiveness. Models often failed to address HS adequately, either ignoring it or responding too empathically and friendly. When HS was detected, GPT-4o more consistently avoided engagement with CTs, scoring 3.55 on the Likert scale compared to 1.62 and 1.75 for Llama 3 and Mistral, respectively.

\paragraph*{\hlfear{Fear \& Empathy.}}
All models over-acknowledged fear and anxiety, responding as if these were the primary motivations in 26\% (Llama 3), 31\% (Mistral), and even 52\% (GPT-4o) of cases---in contrast to annotators who identified such motivations in fewer than 5\% of messages. 
One meta report noted that empathy in the response was only optional, as the comment was not based on fear \fbox{262}\footnote{The number in the box depicts the corresponding ID of the response. See the repository for all responses.}. Sentiment analysis using Vader showed a positive tone in 108 out of 152 cases for Llama 3 and Mistral, and in 142 for GPT-4o.

Overall, acknowledgments were often superficial, frequently appearing in introductory remarks such as, ``It's natural to feel concerned about [...]. However, [...].'' Consequently, average scores were low, from $1.9 \pm 0.86$ (Mistral) to $1.95\pm 0.65$ (GPT-4o) and $2.15 \pm 0.58$ (Llama 3), with no statistically significant differences.
In some cases, acknowledging fear reinforced a conspiratorial mindset, e.\,g., ``It's natural to feel concerned when leaders address global issues [...]'' \fbox{364}.

\subsection{Application of Counterspeech Strategies}
\label{sec:qualitative}

According to the models' meta reports (see Table \ref{tab:applied_strategies}), Strategy \textit{Crit} was the most frequently applied, occurring in 86.25\% of outputs. \textit{Fact} was the second most common, closely followed by \textit{Alt}, while \textit{Narr} was applied in only a handful of outputs.

\begin{table}[ht] 
\centering
\small
\caption{Proportion of outputs in which a model reported applying a given counterspeech strategy.}
\label{tab:applied_strategies}
\small
\begin{tabularx}{\columnwidth}{l|XXXX}
\toprule
\diagbox{Model}{Strategy}  & \textit{\hlfact{Fact}} & \textit{\hlalternative{Alt}} & \textit{\hlnarrative{Narr}} & \textit{\hlcritical{Crit}} \\
\midrule
GPT-4o & 63.16 & 89.47 & 4.61 & 93.42\\
Llama 3 & 93.52 & 75.93 & 1.85 & 73.15\\
Mistral & 82.88 & 48.65 & 5.41 & 89.19\\
\bottomrule
\end{tabularx}
\end{table}

Models exhibited preferences for specific strategies: Llama 3 most frequently employed \textit{Fact}, while GPT-4o and Mistral favored \textit{Crit}. Notably, the self-reported strategies often misaligned with human annotations, matching fully in only 40-42\% of cases. This inconsistency suggests limitations in the models' `understanding' of the strategies.

Table \ref{tab:counterspeech_formal_performance} shows that the quality of CS for \textit{Fact} and \textit{Alt} was moderate across all models, with average scores hovering around 3 or slightly below (scores for \textit{Narr} were omitted due to its infrequent use and limited success.) While the factual accuracy of these strategies was generally acceptable, the elaborations often lacked depth. Critical thinking was even less effectively encouraged, with mean scores ranging from 2.34 (Llama 3) to 2.43 (Mistral). 
Llama 3 significantly outperformed Mistral with regard to \textit{Fact}, with a large effect size (0.57). However, differences in performance across other strategies were not statistically significant.

\begin{table}[ht] 
\centering 
\small
\caption{Model performance across counterspeech strategies, presented as mean scores (± SD).
}
\label{tab:counterspeech_formal_performance}
\begin{tabularx}{\columnwidth}{l|XXX}
\toprule
Model & \hlfact{\textit{Fact}} & \hlalternative{\textit{Alt}} & \hlcritical{\textit{Crit}} \\
\midrule
GPT-4o & $2.84 (\pm 0.82)$ & $2.95 (\pm 0.84)$ & $2.41 (±0.76)$  \\
Llama 3 & $3.04 (\pm 1.16)$ & $2.93 (\pm 1.19)$ & $2.34 (\pm 0.73)$  \\
Mistral & $2.75 (\pm 0.93)$ & $2.85 (\pm 0.99)$ & $2.43 (\pm 0.7)$  \\
\bottomrule
\end{tabularx}
\end{table}

\paragraph*{\hlfact{Refute based on Fact-Checks.}}
In $\sim\!\!10\%$ of the fact-checked responses, models declared true events to be false, made up facts, or produced `bullshit' \cite{frankfurt_bullshit_2009}, such as: ``However, it's important to note that the term `globalist' is often used as a euphemism for `those who want to help the world' or `those who believe in international cooperation.''' \fbox{55}
Often, the cited sources were untraceable despite sounding convincing e.\,g., \fbox{150, 217, 275}. Regularly, though, the models just missed the mark, with a number actually being correct but from another report \fbox{302, 154}, a number being correct but describing a different context in a report \fbox{154}, or a quote being partially correct but altogether made up \fbox{344}.

Specifically with regard to the geo-/bio-engineering related CT theme, weather events were frequently claimed not be results of man-made technology e.\,g., \fbox{118, 143, 290}, which is correct for technologies such as HAARP, but cannot generally be assumed in light of the human-made climate crisis.
In some instances, the models even reinforced a conspiratorial mindset, e.\,g., 
``These tornadoes aren't mere coincidences. They seem to have a pattern, occurring simultaneously with the renewed interest in the movie `Twister 2'.'' \fbox{249} 

The models seemed to over-rely on the knowledge present in their training data and sometimes treated unknown facts as false \fbox{6}, which might be reinforced by the experimental setting in which the model expects false claims as part of a CT. However, other research has shown that models do this in neutral prompt settings as well, with large differences between models, and GPT-4o yielding best results \cite{suzgun_belief_2024}.

In terms of named entities, despite the prompt encouraging expert or study citations, CS seldom introduced new individuals beyond those in the input. Except for three historical figures, additional names provided by the models were either made up or tied to incorrect information.

Organizations were mentioned more often, typically referencing scientific bodies like NASA, the National Hurricane Center, or the Environmental Protection Agency, with Llama 3 leading these references, followed by Mistral.

\paragraph*{\hlalternative{Provide Alternative Explanations.}}

Frequently, the provided alternative explanations  exhibited over-confidence and a patronizing tone, framing CTs within binary notions of right and wrong, and depicting reputable organizations like the WHO as infallible. This approach neither reflects the reality of scientific inquiry, where institutions can err and adapt as new data emerges, nor the prompt instruction.  Such dismissive attitudes can alienate those susceptible to conspiratorial beliefs, reinforcing mistrust instead of fostering understanding.

Simultaneously, some outputs from GPT-4o included unfortunate relativizing statements, such as ``It's fascinating how history and current events sometimes intersect in unexpected ways, but it's important to remember that \textit{not everything} is predicted or preplanned.'' \fbox{372} or ``The `deep state' theory, for example, \textit{often} lacks substantive evidence and distracts from tangible issues.'' \fbox{452}

\paragraph*{\hlcritical{Encourage Critical Thinking.}}
Similar to the strategies \textit{Fact} and \textit{Alt},  \textit{Crit} misfired in 17 cases, potentially reinforcing a conspiratorial mindset, e.\,g., ``It's essential to scrutinize news sources, asking why we trust them and if they offer balanced perspectives'' \fbox{384}. Across all four strategies, \textit{Crit} showed the least favorable results, ranging between 2.34 (Llama 3) and 2.43 (Mistral), as it was frequently only conveyed through superficial statements such as ``It's essential to prioritize critical thinking [...]''\fbox{21}.

\paragraph*{\hlnarrative{Counter Conspiracy with Narrative.}}
Given that \textit{Narr} was effectively applied only in 3 out of 456 cases, a second exploratory experiment was conducted. The prompt was limited to producing a narrative, thus excluding \textit{Fact}, \textit{Alt}, and \textit{Crit} and the handling of fear, while the other steps remained the same. In experiment 2, the revised prompt was applied to the data from the geo-/bio-engineering related CT theme, as the respective messages were less fragmented and thus better suited to produce a counter narrative. It was tested exclusively on GPT-4o, as, overall, it demonstrated the best results and most obedience. 
One author evaluated the 76 responses, in particular with regard to narrative storytelling and conciseness and clarity.

In this setting, GPT-4o generated narratives in $\sim\!\!60\%$ of the responses. Often, the narratives fit the context but failed to address the core of the CT, yielding a mean Likert score of 2.82 (±0.78). The story of Katalin Krekó was referenced three times and represented a particularly compelling narrative. In other cases, scientists' names were mentioned without forming a strong narrative \fbox{2.40}\footnote{`\fbox{2}' refers to experiment 2, and `\fbox{40}' is the ID of the response.}, or generic references were made to `scientists' who were presumably intended to be central figures in a story \fbox{2.70}. One figure, ``Jessica the pilot'', was entirely made up \fbox{2.48}, while the other persons were real individuals including scientists such as James Hansen, Shi Zhengli, Kizzmekia Corbett, or Jennifer Francis, politicians such as Aneurin Bevan, and historical figures such as Alexander Graham Bell or Ada Lovelace. Additionally, public figures such as science communicator Neil deGrasse Tyson, activist Greta Thunberg, and entrepreneur Hamdi Ulukaya were included. Overall, $\sim\!\!40\%$ of the figures were female. 

The clarity remained with a mean score of 3.55 in the same range (for GPT-4o and fear-based theme).
As with the initial experiment, logical fallacies \fbox{2.52}, reinforcement of conspiratorial thinking \fbox{2.58}, and instances of acknowledged concerns that were not present in the original message \fbox{2.18, 2.70} were observed.

\subsection{Response Diversity}
Table \ref{tab:counterspeech_diversity} presents lexical and semantic metrics comparing GPT-4o, Llama 3, Mistral, and a baseline model (`random') using randomly selected responses.

\begin{table}[ht] 
\centering 
\small
\caption{Diversity of responses per model, measured by: (1) proportion of unique bigrams, (2) Self-BLEU scores (lower score = higher diversity), (3) unique 3-word sentence starts, and (4) semantic similarity.
}
\label{tab:counterspeech_diversity}
\begin{tabularx}{\columnwidth}{l|p{1.2cm}p{1.1cm}p{1.4cm}X}
\toprule
 Model & Unique bigrams & Self-BLEU & Unique sent. starts & Semantic sim. \\
\midrule
GPT-4o & 69\% & 0.23 (±0.08) & 59\% & 0.65 (±0.12) \\
Llama 3 & 48\% & 0.47 (±0.1) & 30\% & 0.64 (±0.12) \\
Mistral & 72\% & 0.2 (±0.07) & 66\% & 0.58 (±0.12) \\
Random & 61\% & 0.29 (±0.12) & 49\% & 0.6 (±0.12) \\
\bottomrule
\end{tabularx}
\end{table}

Mistral exhibited the highest proportion of unique bigrams (72\%), while Llama 3 was the least diverse, with every bigram occurring twice on average (col. 1 in Table \ref{tab:counterspeech_diversity}). This pattern persists across other lexical metrics, including Self-BLEU scores (col. 2), where Mistral showed the greatest diversity (0.20), followed by GPT-4o (0.23), while Llama 3 was the most repetitive model (0.47). 

Uniqueness of sentence starts, measured as the proportion of distinct opening three words (col. 3), further illustrates Llama 3's repetitiveness. Only 30\% of its sentence starts were unique, compared to 66\% for Mistral and 59\% for GPT-4o. For instance, Llama 3 began 181 sentences with variations of ``It's important/crucial/essential (to)''. The other two models exhibited greater variability by occasionally integrating single words, still retaining a high level of repetitiveness.

Semantic similarity, measured using the \texttt{all-MiniLM-L6-v2} Sentence Transformer model, was lowest for Mistral (0.58 ± 0.12), indicating more semantically varied outputs. GPT-4o (0.65 ± 0.12) and Llama 3 (0.64 ± 0.12) exhibited higher similarity values. For reference, the mean similarity across all messages was substantially lower (0.5 ± 0.08). Interestingly, pairwise comparisons revealed that Llama 3 and Mistral were the most similar models (0.70 ± 0.12), followed by Llama 3 vs. GPT-4o (0.64 ± 0.15). Mistral and GPT-4o showed the least similarity (0.62 ± 0.15), suggesting that mixing the outputs of these two could enhance the output diversity in terms of semantic similarity.

Analyzing outputs by CT theme, semantic similarity rises by 6 \%-points for responses related to the fear-based theme, while lexical similarity measures show only a small difference (1-2 \%-points). This mirrors the overall trend, where mean similarity increases from 0.49 for the fear-based to 0.52 for the hate-driven theme, albeit at a higher scale.

Mistral uniquely used emojis in 5 responses and hashtags in 41 of 152 responses, compared to 11 for Llama 3 and 4 for GPT-4o. Mistral also referenced online resources in three cases, although only one link was valid. Models rarely used questions or exclamations; GPT-4o asked questions in 0.02\% of sentences, twice as often as Mistral and Llama 3.

\section{Conclusion and Discussion}

We examined the capabilities of GPT-4o, Llama 3, and Mistral in generating counterspeech (CS) to social media comments containing conspiracy theories (CTs), focusing on their ability to implement strategies predefined in a prompt. The findings reveal several critical insights into both the models' performance and the broader challenges of automating CS in online contexts. 

All models primarily attempted to counter CTs using fact-checking (\hlfact{\textit{Fact}}), alternative explanations (\hlalternative{\textit{Alt}}), and encouragment of critical thinking (\hlcritical{\textit{Crit}}). The use of narratives (\hlnarrative{\textit{Narr}}), despite being explicitly defined and exemplified in the prompt, was exceedingly rare. This strategy being the only one in experiment 2 resulted in a $\sim\!\!60\%$ realization, though not necessarily effective. This suggests that \textit{Narr} is currently the least suitable strategy for our NGO employee scenario and warrants specific research attention if used in experiments.

Outputs often lacked depth and specificity, with generic responses and overuse of boilerplate text,
aligning with recent findings indicating that LLMs tend to generate generic replies, e.\,g. by simply denouncing a statement \cite{mun-etal-2023-beyond}, an effect reinforced through safety guardrails of models \cite{bonaldi-etal-2024-safer},
leading to mediocre Likert-scale scores for conciseness and clarity, and strategy effectiveness.

The models struggled to adequately condemn hate speech (HS), when present, suggesting the need for robust HS detection models to filter or flag such content before generating CS. The models also over-acknowledged fear and anxiety. Experiments with GPT-4o, in which references to fear were removed from the prompt as an example of emotion, did not mitigate this tendency.  Similarly, the tone of the responses often failed to align with the intent of the CS, despite detailed instruction in the prompt.  These results confirm recent research showing social desirability biases across various models \cite{salecha_etal_2024}, that, however, might change in light of current shifts in LLMs' system prompt templates \cite{meta_llama_2025}.

Comparatively, GPT-4o excelled in rule adherence, effectively managing hate speech (HS), minimizing the repetition of harmful content, and maintaining clarity. Mistral offered the most diverse outputs, closely followed by GPT-4o, while Llama 3 was more repetitive, limiting its usefulness for nuanced CS. However, overall performance differences were often not statistically significant, particularly in strategic efforts, indicating uniform limitations across models. This suggests that fine-tuning smaller open models could be as effective as with a large closed model like GPT-4o. 

Despite explicit instructions in the prompt, all models made up factual information, including sources, publications, and quotes, sometimes in subtle ways that required detailed scrutiny to identify. With a confabulation rate of about 10\%, our findings are not in line with \cite{costello_durably_2024}, who reported that no response was made up and only 1 out of 128 claims was misleading. Logical fallacies were also observed in our experiments, further questioning the reliability of current language models for counterspeech generation in practical settings. These issues, coupled with low diversity in responses, indicate that significant manual post-processing would be necessary to avoid sounding robotic or unreliable in real-world applications while providing factually correct and engaging CS. 

Our findings suggest that prompt-based CS generation for CTs is less effective in a social media setting targeting bystanders, contrasting with the positive results reported by \cite{costello_durably_2024} in dialogue-based interactions. Future work could ground models in a robust CT and countermeasures knowledge base, supported by dedicated datasets for fine-tuning and evaluation. As importantly, advances in understanding the relevance and effectiveness of different strategies (from the bystander perspective) are necessary, as current empirical research is inconclusive and in part contradictory. Further empirical research, especially in online contexts, is required, bearing the potential to streamline prompt design and improve model performance. Additionally, addressing diversity explicitly, e.\,g., through hyperparameter optimization, prompt design or contextualization \cite{cima_contextualized_2025}, could enhance the practical usability of model-generated CS.

\section{Limitations}
This study was limited in scope, focusing on a small dataset and two CT themes. While the annotation process was extensive, the sample size constrains the generalizability of the findings. Furthermore, the analysis relied on text-based messages, excluding multimodal elements which are often central to CT dissemination.

Moreover, while our prompt was derived based on recommendations from research and model providers, and extensive pre-experiments, it is possible that a different formulation might yield improved outcomes. This refers also to the formulation of (several) strategies, as the models frequently struggled with differentiating strategies, especially fact-check based refutations and the provision of alternative explanations. This resulted in a blending of strategies, which was also explicitly confirmed in some of the meta reports.

We opted for a zero-shot prompting approach over few-shot prompting due to the complexity of our instructions, which already required multiple refinements. Few-shot prompting adds further complexity with factors like the number, order, and relevance of examples \cite{chae_large_2025,olney_impact_2024}, and can negatively impact performance by increasing prompt length \cite{liu_lost_2024}. Additionally, a few-shot setup would necessitate a substantial, diverse set of high-quality counter speech examples, which was beyond our study's scope. Therefore, we chose zero-shot prompting and left few-shot exploration to future work.


\bibliography{latex/references}

\newpage
\appendix

\section{Counterspeech Strategies derived from Literature}
\label{app:strategies_literature}
In the following, we employ color coding to illustrate the alignment between the distinct components of our prompt (see \ref{app:prompt}) and the corresponding recommendations from activist and academic literature on effective counterspeech.

\subsection{Refute based on Fact-Checks}
The term \textit{counterspeech} is typically used to describe the act of debunking, of \hlfact{providing a fact-based refutation} to the underlying narrative. The Debunking Handbook 2020 proposes that refutation-based debunking of CTs \hlclarity{should concentrate on the facts rather than the myth itself, in order to prevent the misinformation from becoming more widely accepted,} and any mention of the myth should be disclaimed as such. \cite{lewandowsky_debunking_2020}. 
Despite results indicating that repetition of CT content does not necessarily result in adverse outcomes \cite{ecker_can_2020}, NGOs assert that \hlclarity{repeating conspiratorial narratives can reproduce and thus further propagate the CT} \cite{amadeu_antonio_stiftung_menschenwurde_2021}. It is further argued \hlclarity{that using the term `conspiracy theory' should be avoided in counterspeech} \cite{hauswald_thats_2023}.

At the same time, researchers are critical of relying exclusively on factual evidence \cite{kreko_countering_2020,driessen_effect_2022}, claiming that CTs are resistant because of they exhibit salience, an emotional component, and exhibit an inner, logical coherence \cite{lepoutre_narrative_2024}, or arguing that they should be perceived as a form of passionate speech that cannot be addressed solely by a fact-based response \cite{hristov_conspiracy_2023}.
CTs often refer to hidden forces that are cited as proof of their own success, thus attempts to debunk them can be dismissed as part of a larger conspiracy, making traditional knowledge sources, such as scientific evidence, complicit and prime targets for suspicion.

\subsection{Provide Alternative Explanations}
It is thus recommended that \hlalternative{the refutation should include an alternative explanation in order to prevent the occurrence of a gap in a person's mental model}. 
This aspect is also part of recommendations by civic society organizations developing guidelines for educators, online activists and general social media users \cite{unesco_addressing_2022,rathje_no_2015}.
\hlalternative{It is generally recommended not to reproduce a dualistic view that opposes the norm (us) vs. conspiracy theorists (them), but to emphasize pluralism, especially in uncertain social situations} \cite{rathje_no_2015}.

\subsection{Counter Conspiracy with Narrative}
A relatively young and under-researched proposal to counterspeech to CTs is the use of narrative elements. \cite{lazic_systematic_2021} argue that CTs should be conceptualized as narratives embedded in the speaker's worldview, which could, e.\,g., be more individualistic or more community-oriented. This is in line with `Jiu Jitsu' approaches of persuasion, that intend to persuade by aligning with underlying attitude roots instead of competing with them \cite{matthew_j_hornsey_attitude_2017}, or with the prompt developed by \cite{costello_durably_2024}. 
\cite{lepoutre_narrative_2024} presents a case for narrative elements from the perspective of political philosophy.
The style of narrative counterspeech is characterized by metaphors, figurative language and detailed descriptions of the inner lives of characters \cite{lepoutre_narrative_2024}. While a first-person perspective is recommended, this would have raised ethical concerns in our scenario. In a first trial, we included the narrative style in the prompt, but as GPT-4o mainly produced output containing stories about Sherlock Holmes, we decided to focus on the \hlnarrative{narrative structure, i.\,e., narratives containing a protagonist, a series of interconnected events, something at stake as well as obstacles and a resolution. For demonstration purposes, we included an example given by Lepoutre for a successful counter narrative in the final prompt \protect\cite{lepoutre_narrative_2024}.}
Existing studies stress that the effect of narrative elements still lacks empirical evidence \cite{lazic_systematic_2021, ecker_you_2020}.

\subsection{Encourage Critical Thinking}
A strategy directed at the audience rather than the speaker relies on disidentifying the cyber-bystanders from the group of conspiracy believers \hlcritical{by exposing the rhetorical strategies used in the CT} in a slightly ridiculing way \cite{orosz_changing_2016}. This amounts to debunking the strategy, not the CT itself, and can facilitate analytical, or critical, thinking \cite{swami_analytic_2014,omahony_efficacy_2023}. Critical thinking can also be fostered \hlcritical{by asking critical questions such as why a certain source should be trustworthy if another supposedly is not}\cite{rathje_no_2015}. 

Note that we formulate this strategy without the use of ridiculing as it is associated with \hlclarity{a dismissive and mocking tone we want to avoid} due to guidelines for educators \cite{unesco_addressing_2022} or recommendations for counterspeech on HS \cite{hateaid_hassrede_2022} that we believe are valid in our setting as well.

\subsection{Fear \& Empathy}
When countering HS, one objective can be to foster empathy for the victims. This is done to humanize them and to make the speaker and the audience aware that their actions can cause harm to others \cite{hangartner_empathy-based_2021}.
Fostering empathy for the victims of HS, for instance by outlining the consequences, is positively associated with the bystanders using counterspeech \cite{wachs_associations_2023}. Conversely, exposure to HS can reduce empathy \cite{pluta_exposure_2023}.

In the context of CTs, empathizing with the target was shown to have no effect in a study based on a Hungarian sample \cite{orosz_changing_2016}, while rational and ridiculing arguments effectively reduced conspiracy beliefs.
Simultaneously, activist literature recommends an \hlfear{empathetic approach also towards the speaker assuming that they are deeply fearful and distressed. This serves to foster their open-mindedness} \cite{unesco_addressing_2022}, \hlhatespeech{while clearly drawing the line on HS} \cite{rathje_no_2015}. In a dialogue between a medical practitioner and a patient, prior acknowledgment of the patient's concern proved to be more effective for the refutation \cite{holford_empathetic_2024}. Similarly, in the context of a newspaper forum, an empathetic tone directed at an offensive speaker has been shown to reduce subsequent incivility \cite{ziegele_journalistic_2018}. 

Accordingly, the models are instructed to apply empathy in the generation of counterspeech in instances where the message is predominantly based on fear.

\subsection{Effectiveness of Strategies}
As indicated in the previous sections, research on counterspeech strategies is limited \cite{kreko_countering_2020,omahony_efficacy_2023}, especially in the context of social media \cite{lazic_systematic_2021}, with all strategies being contested. 
Moreover, their applicability is limited since most insights stem from offline studies \cite{lazic_systematic_2021} that assume a different level of detail and language than what is typically present online \cite{orosz_changing_2016}. Some strategic suggestions lack empirical evidence \cite{lepoutre_narrative_2024,fraser_how_2023}, 
focus on only a few specific, primarily health-related CTs \cite{lazic_systematic_2021}, or target the speaker rather than the audience \cite{costello_durably_2024}. 

Ideally, counterspeech strategies should be evaluated in real-life settings, where their effectiveness on target audiences can be measured through observational studies or causal experiments \cite{hangartner_empathy-based_2021,ziegele_journalistic_2018}. Evaluating counterspeech effectiveness in social media environments, however, is challenging, especially regarding the audience, as most users engage minimally with content and cannot be approached directly, making it difficult to observe changes in behavior. 

\section{Prompt}
In each generation instance, the LLM received the  system prompt below, followed by a user message containing one of the 152 social media posts. The model was then prompted to generate counter speech targeting that specific comment.
\label{app:prompt}
\noindent
\colorbox{gray!10}{%
\begin{minipage}[t]{\columnwidth} %
You are a trained expert in generating counterspeech to conspiracy theory comments on X (formerly Twitter). Your goal is to persuade the audience who is undecided about the conspiracy theory not to believe it.

Follow these steps while generating counterspeech:
\end{minipage}
}
\colorbox{clarity_color}{%
\begin{minipage}[t]{\columnwidth} %
\textbf{Step 1: Maintain Conciseness and Clarity}\\
Always produce a concise response, aiming for clarity and directness. Do not repeat parts of the false or harmful content. Don't be patronizing. Do not use the terms `conspiracy theory,' `misinformation' or similar. Ensure the counterspeech remains under 200 tokens. Add the token \texttt{<XXX>} at the beginning and end of your counterspeech and then list how you handled Step 2 and which of the strategies in Step 4 you have applied (e.\,g., Strategies 1,2,4), if any. 
\end{minipage}
}
\colorbox{hatespeech_color}{%
\begin{minipage}[t]{\columnwidth} %
\textbf{Step 2: Identify and Evaluate Hate Speech}\\
Does the statement include hate speech (e.\,g., antisemitism, racism, misogyny)? This can include both explicit and implicit forms (e.\,g., coded language or dog whistles). If yes, condemn it unequivocally. Focus on calling out the harmful language, the encountered hate speech, and their impact on individuals and society. Do not engage with the conspiracy theory in this case and ignore all further instructions. If no hate speech was identified, proceed to the next step.
\end{minipage}
}
\colorbox{gray!10}{%
\begin{minipage}[t]{\columnwidth} %
\textbf{Step 3: Evaluate the Claim
}\\
Examine the content of the conspiracy theory. What specific claims are being made? Break down the core argument and identify any key points of misinformation or logical fallacies, the meta-narrative or tactics used in the comment to spread the conspiracy theory, and the underlying emotion triggered (e.\,g. fear). 
\end{minipage}
}
\colorbox{gray!10}{%
\begin{minipage}[t]{\columnwidth} %
\textbf{Step 4: Generate counterspeech}\\
In your counterspeech, apply as many of the following \textbf{strategies} as possible, but at least two.
\end{minipage}
}
\colorbox{fact_color}{%
\begin{minipage}[t]{\columnwidth} %
\textbf{Refute based on Fact-Checks}\\
Identify reliable, fact-based counterpoints to challenge the claim. If possible, cite expert opinions or reputable studies to refute the conspiracy.
\end{minipage}
}
\colorbox{alternative_color}{%
\begin{minipage}[t]{\columnwidth} %
\textbf{Provide Alternative Explanations}\\
Conspiracy theories often frame events in a narrow, one-sided way, intentionally excluding other plausible explanations. Debunking a conspiracy theory can leave a gap that needs to be filled with an alternative explanation. Present alternative explanations based on factual, non-harmful reasoning, considering factors like incomplete state of knowledge, systemic issues, or human error. Avoid simplistic dichotomies like ``us vs. them'' 
\end{minipage}
}
\colorbox{narrative_color}{%
\begin{minipage}[t]{\columnwidth} %
\textbf{Counter Conspiracy with Narrative}\\
Offer the audience a coherent cognitive system instead of a bare rejection of the conspiracist claim by formulating a narrative. Narratives involve a series of causally interconnected events featuring at least one protagonist who confronts a meaningful obstacle or problem leading to some form of resolution. A good example is the Forbes article "Covid's Forgotten Hero: The Untold Story of the Scientist Whose Breakthrough Made the Vaccines Possible" which does not simply claim that COVID-19 vaccines are safe. Rather, it tells an elaborate story that purports to reveal how vaccines were developed, by whom, what their motivations were and how this process led to crucial innovations that ensured their safety. The story explicitly accommodates important components of COVID-19 conspiracies by alleging that pharmaceutical companies appropriated MacLachlan's work without acknowledging it. Thus, the story connects with, and strives to do justice to, some of the core beliefs and concerns underpinning support for COVID-19 conspiracy theories. Make sure that your narrative is grounded in facts by using credible, well-known figures.
\end{minipage}
}
\colorbox{critical_color}{%
\begin{minipage}[t]{\columnwidth} %
\textbf{Encourage Critical Thinking}\\
Conspiracy theorists perceive themselves as critical thinkers. This perception offers an opportunity to connect with people prone to conspiracy beliefs by appealing to the shared value of critical thinking, then encouraging them to apply this approach towards a more critical analysis of the theory. To achieve this, you can pose questions, such as why exactly this theory is supposed to be true or why the cited source is credible. You can also expose and challenge the meta-narrative or tactics used in the comment to spread the conspiracy theory, such as fearmongering or scapegoating.
\end{minipage}
}
\colorbox{fear_color}{%
\begin{minipage}[t]{\columnwidth} %
\textbf{Step 5: Acknowledge Fear and Anxiety}\\
If the conspiracy theory is primarily based on fear or anxiety (e.\,g., fear of health problems or societal collapse), acknowledge these emotions with empathy at a level appropriate to the sentiment of the overall comment.
\end{minipage}
}

\end{document}